\documentclass[11pt]{article}
\usepackage[dvipsnames]{xcolor}
\usepackage[nohyperref, final]{acl}
\usepackage{times}
\usepackage[most]{tcolorbox}
\usepackage{courier}
\usepackage{latexsym}
\usepackage[normalem]{ulem}
\usepackage[T1]{fontenc}
\usepackage[utf8]{inputenc}
\usepackage{microtype}
\usepackage{subfiles}
\usepackage{inconsolata}
\usepackage{graphicx}
\usepackage{amsthm}
\usepackage{adjustbox}
\usepackage{url}    
\usepackage{bussproofs}
\usepackage{ragged2e}
\usepackage{soul}
\usepackage{booktabs}   
\usepackage{algcompatible}
\usepackage{subcaption}
\usepackage{amsfonts}     
\usepackage{nicefrac}   
\usepackage{svg}
\usepackage{titlesec}
\usepackage{tikz}
\usepackage{amssymb}
\usepackage{amsmath}
\usepackage[capbesideposition=right]{floatrow}
\usepackage{float}
\usepackage[inline]{enumitem}
\usepackage{wrapfig}
\usepackage{prftree}
\usepackage{tabularx}
\usepackage{colortbl}
\usepackage{hhline}
\usepackage{pifont}
\usepackage{bm}
\usepackage[ruled,linesnumbered]{algorithm2e}
\SetAlgoNlRelativeSize{-1}
\usepackage{dsfont}
\usepackage{setspace}
\usepackage[
    colorlinks=true,
    linkcolor={red!55!black},
    citecolor={blue!65!black},
    urlcolor={blue!65!black},
    hyperfootnotes=true
]{hyperref}
\definecolor{pale_green}{rgb}{0.55,0.75,0.60}
\definecolor{pale_red}{rgb}{0.90,0.61,0.58}
\definecolor{pale_yellow}{rgb}{0.95,0.92,0.72}

\newcommand{\bluett}[1]{\texttt{\textcolor{RoyalBlue}{#1}}}
\newcommand{\cmark}{\ding{51}}
\newcommand{\hlc}[2][yellow]{{%
    \colorlet{foo}{#1}%
    \sethlcolor{foo}\hl{#2}}%
}

\newcommand{\dataset}{\textsc{InAbHyD}}
\newcommand{\xmark}{\ding{55}}

\newcommand{\inlinecode}[1]{%
  \texttt{%
    \setlength{\spaceskip}{0.5em plus 1em minus 2em}%
    \ifdim\lastskip>0pt \unskip\hspace{0.5em plus 0.5em minus 0.1em}\fi
    #1
  }%
}

\theoremstyle{definition}
\usetikzlibrary{decorations.pathreplacing,calligraphy,positioning,shapes.multipart,arrows,shapes.geometric,arrows.meta}

\title{Do Language Models Follow Occam’s Razor? \\An Evaluation of Parsimony in Inductive and Abductive Reasoning}
\author{Yunxin Sun  \and Abulhair Saparov\\
       \texttt{ \{sun1114, asaparov\}@purdue.edu} \\ Purdue University}

\newif\ifshowrevision
\showrevisionfalse
\definecolor{revision}{RGB}{0,128,128}

\ifshowrevision
    \newcommand{\rev}[1]{\textcolor{revision}{#1}}
    \newcommand{\old}[1]{\textcolor{red}{\sout{#1}}}
\else
    \newcommand{\rev}[1]{#1}
    \newcommand{\old}[1]{}
\fi
\begin{document}
\maketitle
\begin{abstract}

Non-deductive reasoning, encompassing inductive and abductive reasoning, is essential in addressing complex real-world questions.  One key feature of inductive and abductive reasoning is that there are many valid hypotheses; the simplest ones (those that adhere to Occam's Razor) are often most useful. However, this aspect is ignored in recent work that evaluates the non-deductive reasoning capabilities of large language models (LLMs). This work fills this gap, focusing on understanding whether the inductive and abductive reasoning capabilities of LLMs adhere to Occam's Razor, while also examining the correctness of their reasoning.  To accomplish this goal, we introduce a framework to synthetically generate reasoning questions that \emph{(a)} require inductive reasoning and abductive reasoning simultaneously;  \emph{(b)} 
is readily extended to produce any abductive/inductive reasoning question expressible in first-order logic.  The task for the intelligent agent is to produce hypotheses to \emph{explain} observations under a given world model. We also propose a new automated metric to assess whether hypotheses quantitatively adhere to Occam's Razor; those hypotheses that are correct and simplest are considered \emph{high-quality}. Our findings on state-of-the-art LLMs suggest that LLMs can perform inductive and abductive reasoning in simple scenarios, but struggle with complex world models and with producing \emph{high-quality} hypotheses, even with popular reasoning-enhancing techniques such as in-context learning and RLVR.
\footnote{All data and generation code are freely available at: \\ \phantom{aaaa}\url{https://github.com/byrantwithyou/inabhyd}}

\end{abstract}

\section{Introduction}
Reasoning is a hallmark of intelligence. Peirce categorizes reasoning into deductive, inductive, and abductive reasoning ~\citep{peirce1878deduction,cat}, all of which are indispensable in real-world scenarios. Deductive reasoning draws a conclusion from a set of premises.  
Inductive reasoning aims to identify a general principle from multiple individual cases. Abductive reasoning seeks to produce hypotheses to explain observations.
\rev{See Appendix~\ref{sec:prelim} for examples of different types of reasoning and for further background}.

\begin{figure*}
\centering
    \begin{subfigure}[b]{1\textwidth}
        \centering 
    \includegraphics[width=1\textwidth]{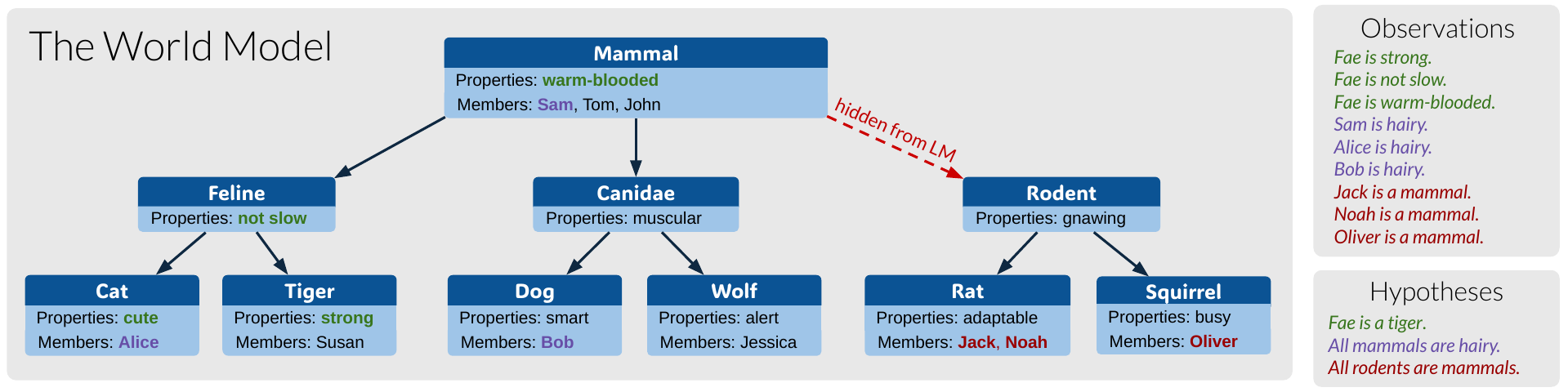}
    \end{subfigure}
    \begin{subfigure}[b]{1\textwidth}
        \centering
    \begin{tikzpicture}
        \definecolor{color2}{RGB}{52, 138, 189}
        \definecolor{color3}{RGB}{152, 142, 213}
        \definecolor{color4}{RGB}{251, 193, 94}
        \definecolor{color5}{RGB}{142, 186, 66}
        \node[text width=35em ] (question) {
            \begin{spacing}{1.0}
           \noindent\justifying\footnotesize\textit{\hlc[color4]{\textbf{Q:} All mammals are warm-blooded. Sam is a mammal. Tom is a mammal. John is a mammal. Each feline is a mammal. All canidae are mammals. All canidae are muscular. Every dog is a canidae. All wolves are canidae. All felines are not slow. 
            Each cat is a feline.
             All tigers are felines. All cats are cute.
            Alice is a cat. All dogs are smart. Bob is a dog. All wolves are gregarious. Jessica is a wolf. All tigers are strong. Susan is a tiger. All rodents are gnawing. Each rat is a rodent. Each squirrel is a rodent. All rats are adaptable. Jack is a rat. Noah is a rat. All squirrels are bushy. Oliver is a squirrel.} \hlc[color2!70]{We observe that Sam is hairy. Alice is hairy. Bob is hairy. Fae is strong. Fae is not slow. Fae is warm-blooded. Jack is a mammal. Noah is a mammal. Oliver is a mammal.}} \newline \textit{Please produce hypotheses to explain all observations.} \newline 
            \textit{\hlc[color5!90]{\textbf{A:} All mammals are hairy. Fae is a tiger. All rodents are mammals.} }
            \end{spacing}
        };
        \node[right=1.2em of question,yshift=2.5em] (context) { \color{black!70}\footnotesize\hlc[color4]{\textsf{world model}} };
        \node[right=1.2em of question,yshift=-0.6em] (query) { \color{black!70}\footnotesize\hlc[color2!70]{\textsf{observations}} };
        \node[right=1.2em of question,yshift=-3.5em] (cot) { \color{black!70}\footnotesize\hlc[color5!90]{\textsf{hypotheses}} };
        \draw[draw=black!70] (context.west) -- ([xshift=-1.2em]context.west);
        \draw[draw=black!70] (query.west) -- ([xshift=-1.2em]query.west);
        \draw[draw=black!70] (cot.west) -- ([xshift=-1.2em]cot.west);
    \end{tikzpicture}
    \end{subfigure}
    \caption{An example of a question requiring inductive and abductive reasoning. Note that the dashed arrow means the subtype relationship \bluett{All rats are mammals} is hidden and needs to be inferred (i.e., a hypothesis to propose).  This example uses a real-world model for illustrative purposes. In our dataset, we use fictional world models.}
    \label{fig:banner}
\end{figure*}

There is a significant body of work on evaluating the deductive reasoning capabilities of LLMs ~\citep{prontoqa,prontoood,logiqa,proofwriter,follo}, as well as techniques to improve deductive reasoning capabilities, such as in-context learning and reinforcement learning with verifiable rewards, \emph{RLVR} ~\citep{tulu, RLVRmultiple}. However, there is relatively little exploration into the inductive and abductive reasoning capabilities of large language models (LLMs). A fundamental difference between deductive and inductive or abductive reasoning is that deductive reasoning is deterministic. However, inductive and abductive reasoning are inherently probabilistic. There may exist many hypotheses, some \emph{more probable} and some \emph{less probable}.  For example, if we hear a bird singing, a \emph{more probable} hypothesis is that there is a bird outside; while a \emph{less probable} hypothesis is that someone outside is playing the sound on a speaker. 
Among these possible hypotheses, the simplest ones are often more useful, and we view them as more probable. 
This principle is known as \emph{Occam's Razor}. We call hypotheses that adhere to Occam's Razor \emph{high-quality} hypotheses. 
Knowing and applying Occam’s Razor is essential in areas such as scientific discovery, which require inductive and abductive reasoning, yet this is often overlooked. Physicists, for example, strive for the simplest explanations to account for as many observations as possible. 


The high-level research question we aim to answer is: \emph{Do LLMs reason inductively and abductively? If so, how well do they adhere to Occam's Razor?} This work systematically evaluates whether and to what extent LLMs can reason inductively and abductively and adhere to Occam's Razor. We also assess whether popular reasoning-enhancing techniques, such as in-context learning ~\citep{ICL1, ICL2} and RLVR ~\citep{tulu, RLVRmultiple}, which primarily aim to improve deductive reasoning capabilities, can also improve inductive and abductive reasoning.

To answer this research question, we introduce a framework to synthetically generate reasoning questions that \emph{(a)} contain inductive reasoning and abductive reasoning
simultaneously; \emph{(b)} is readily extended to produce any inductive/abductive reasoning question expressible in first-order logic. 
 Each reasoning example has a fictional \emph{incomplete} world model in the form of an ontology tree and a set of observations, represented in first-order logic. To answer each reasoning example, LLMs must produce hypotheses to explain observations under the incomplete world model. We generate synthetic examples of fictional world models to mitigate the contamination of training data. The ontology tree's height characterizes the difficulty of the world model and the corresponding reasoning example.\footnote{Refer to Figure~\ref{fig:banner} and Appendix~\ref{sec:more_examples} for examples.} To assess whether LLMs adhere to Occam's Razor quantitatively, we also introduce an automated metric to quantify the quality of each hypothesis (i.e., \emph{how well each hypothesis adheres to Occam's Razor}).
We also introduce a new benchmark dataset, \dataset (\textbf{In}ductive and \textbf{Ab}ductive \textbf{Hy}pothesis \textbf{D}iscovery), generated under this framework. This dataset contains over 2K reasoning questions with different difficulty parameters. \rev{The dataset also includes the generation code, which can, in principle, be used to generate an unbounded number of reasoning questions.}

We evaluate our dataset on  \old{five} \rev{seven} widely-used LLMs: \rev{\textsc{GPT-5.4}}~\citep{gpt5}, \textsc{\rev{o3}}~\citep{o3}, \textsc{Llama3-70B} ~\citep{llama} , \textsc{Gemma3-27B} ~\citep{gemma}, \textsc{DeepSeek-V3}  ~\citep{v3},  \textsc{GPT-4o} ~\citep{gpt4} , and \textsc{DeepSeek-R1-Distill-Llama-70B} ~\citep{r1}. For each LLM, we test whether it can produce \emph{valid} and \emph{high-quality} hypotheses. We also test whether popular reasoning-enhancing techniques, such as in-context learning and RLVR, which primarily were meant to improve deductive reasoning, can also improve inductive and abductive reasoning. Our findings can be summarized as follows:
\begin{enumerate}
    \item LLMs can perform inductive and abductive reasoning at high accuracy for shallow ontology trees with a single hypothesis.
    \item The accuracy drops significantly when the ontology tree structure is complex and contains multiple hypotheses.
    \item In most cases, LLMs may produce \emph{valid} hypotheses (which can explain all observations); but fail to produce \emph{high-quality} hypotheses (which adhere to Occam's Razor).
    \item In general, we observe that reasoning-enhancing techniques \emph{moderately} improves inductive and abductive reasoning. 
\end{enumerate}

\section{Related Work}
\paragraph{Evaluating non-deductive reasoning capabilities.}
\begin{table*}
	\small
	\sffamily
	\def\arraystretch{1.15}
	\setlength{\tabcolsep}{2pt}
\begin{tabular*}{\textwidth}{m{0.28\textwidth}>{\centering\arraybackslash}m{0.12\textwidth}>{\centering\arraybackslash}m{0.11\textwidth}>{\centering\arraybackslash}m{0.10\textwidth}>{\centering\arraybackslash}m{0.10\textwidth}>{\centering\arraybackslash}m{0.12\textwidth}>{\centering\arraybackslash}m{0.09\textwidth}}
		\hhline{|-|------|}
		\centering\arraybackslash\textbf{Dataset} & Controls for complexity of world models & Natural language description & Tests for inductive reasoning & Test for abductive reasoning & Evaluation of hypotheses quality & \hspace{0.5em}Multiple \newline hypotheses \\
		\hhline{:=:======:}
		\texttt{ART} {\scriptsize\citep{anll}} &
			\cellcolor{pale_red!35} \xmark &
			\cellcolor{pale_green!35} \cmark &
			\cellcolor{pale_red!35} \xmark &
			\cellcolor{pale_green!35} \cmark &
			\cellcolor{pale_yellow!35} {\scriptsize non-quantitative} &
			\cellcolor{pale_red!35} \xmark \\
		\hhline{|-|------|}
		\texttt{InductiveBench} {\scriptsize\citep{inductivebench}} &
			\cellcolor{pale_yellow!35} N\textbackslash A &
			\cellcolor{pale_red!35} \xmark &
			\cellcolor{pale_green!35} \cmark &
			\cellcolor{pale_red!35} \xmark &
			\cellcolor{pale_red!35} \xmark &
			\cellcolor{pale_red!35} \xmark \\
		\hhline{|-|------|}
		\texttt{AbductionRules} {\scriptsize\citep{abductionrule}} &
			\cellcolor{pale_red!35} \xmark &
			\cellcolor{pale_green!35} \cmark &
			\cellcolor{pale_red!35} \xmark &
			\cellcolor{pale_green!35} \cmark &
			\cellcolor{pale_red!35} \xmark &
			\cellcolor{pale_red!35} \xmark \\
		\hhline{|-|------|}
		\texttt{Wino-Why} {\scriptsize\citep{winowhy}} &
			\cellcolor{pale_red!35} \xmark &
			\cellcolor{pale_green!35} \cmark &
			\cellcolor{pale_red!35} \xmark &
			\cellcolor{pale_green!35} \cmark &
			\cellcolor{pale_red!35} \xmark &
			\cellcolor{pale_red!35} \xmark \\
        \hhline{|-|------|}
		\texttt{SyGus} {\scriptsize\citep{sygus}} &
			\cellcolor{pale_yellow!35} N\textbackslash A &
			\cellcolor{pale_red!35} \xmark &
			\cellcolor{pale_green!35} \cmark &
			\cellcolor{pale_red!35} \xmark &
			\cellcolor{pale_red!35} \xmark &
			\cellcolor{pale_red!35} \xmark \\
        \hhline{|-|------|}
		\texttt{UniADILR} {\scriptsize\citep{Uni}} &
			\cellcolor{pale_red!35} \xmark &
			\cellcolor{pale_green!35} \cmark &
			\cellcolor{pale_green!35} \cmark &
			\cellcolor{pale_green!35} \cmark &
			\cellcolor{pale_red!35} \xmark &
			\cellcolor{pale_red!35} \xmark \\
        \hhline{|-|------|}
		\texttt{LogiEval} {\scriptsize\citep{logieval}} &
			\cellcolor{pale_red!35} \xmark &
			\cellcolor{pale_green!35} \cmark &
			\cellcolor{pale_green!35} \cmark &
			\cellcolor{pale_green!35} \cmark &
			\cellcolor{pale_red!35} \xmark &
			\cellcolor{pale_red!35} \xmark \\
        \hhline{|-|------|}
		\texttt{MME-Reasoning} {\scriptsize\citep{mme_reasoning}} &
			\cellcolor{pale_red!35} \xmark &
			\cellcolor{pale_green!35} \cmark &
			\cellcolor{pale_green!35} \cmark &
			\cellcolor{pale_green!35} \cmark &
			\cellcolor{pale_red!35} \xmark &
			\cellcolor{pale_red!35} \xmark \\
    
		\hhline{:=:======:}
		  \dataset(\textbf{ours}) &
			\cellcolor{pale_green!35} \cmark &
			\cellcolor{pale_green!35} \cmark &
			\cellcolor{pale_green!35} \cmark &
			\cellcolor{pale_green!35} \cmark &
			\cellcolor{pale_green!35} \cmark &
			\cellcolor{pale_green!35} \cmark \\
		\hhline{|-|------|}
	\end{tabular*}

	\caption{Comparison of existing datasets to evaluate inductive and abductive reasoning capabilities.
    }\label{tab:datasets}
\end{table*}

Table~\ref{tab:datasets} summarizes the similarities and differences between our work and others on evaluating non-deductive reasoning capabilities of both LLMs and earlier models.
\rev{Representative works on abductive reasoning evaluation, such as ART~\citep{anll}, AbductionRules~\citep{abductionrule}, and Wino-Why~\citep{winowhy}, craft abductive reasoning questions from web sources, and ask LMs to explain everyday events. However, this approach confounds the problem of commonsense knowledge acquisition with the problem of abductive reasoning. In contrast, our framework provides all required facts in the context of each prompt, thereby evaluating LM's abductive reasoning capabilities in a more controlled fashion.}
\rev{Representative works on inductive reasoning evaluation, such as InductiveBench~\citep{inductivebench} and SyGus~\citep{sygus} focus exclusively on highly-structured domains such as string translation and program synthesis. However, natural language-based inductive reasoning is also essential and is underexplored.}

\paragraph{Evaluating deductive reasoning capabilities.}
There is no consensus on whether LLMs can perform general deductive reasoning or are stochastic parrots~\citep{parrot}. Researchers have designed different datasets and evaluation methodologies to answer this question ~\citep{prontoood, prontoqa, logiqa, proofwriter, follo}. \textsc{LogiQA} ~\citep{logiqa} is a multiple-choice QA dataset to test multiple types of deductive reasoning of language models. However, it focuses solely on label accuracy (\emph{the correctness of the final answer}) without considering the correctness of intermediate proof steps. In contrast, \textsc{PrOntoQA} ~\citep{prontoqa,prontoood} evaluates both label accuracy and step-by-step proof correctness.  They can synthetically generate proofs with different heights, widths, and deduction rules.  
Our work is similar to \textsc{PrOntoQA} in spirit. We both use a
first-order ontology tree as the world model. However, \textsc{PrOntoQA} does not evaluate non-deductive reasoning.

\paragraph{Techniques to enhance reasoning.}
A large body of work has focused on improving the 
 (deductive) reasoning capabilities of LLMs and related downstream tasks such as mathematics and coding ~\citep{ttt1, CoT, ToT, consistency, RLVRmultiple, tulu, ICL1, ICL2}.  The first line of work aims to elicit the emergent capabilities of LLMs using different prompt formats, such as chain-of-thought ~\citep[CoT;][]{CoT}, tree-of-thought ~\citep[ToT;][]{ToT}, self-consistency ~\citep{consistency}, and in-context learning.  The second line of work post-trains LLMs with reinforcement learning with verifiable reward (\emph{RLVR}; ~\citealt{tulu, RLVRmultiple}) to improve LLM reasoning. \rev{Since non-deductive reasoning is inherently ambiguous, it is difficult to define a verifiable reward in non-deductive settings.
 The third line of work uses neuro-symbolic approaches to enhance the efficiency and reliability of LLM reasoning~\citep {ns1,ns2,ns3}. However, they are either not directly applicable to unstructured natural language input or designed for deductive reasoning applications.}
 This work investigates whether the above methods benefit inductive and abductive reasoning.

\section{Approach}

\subsection{Overview}
In this section, we describe how we generate the dataset and evaluate LLMs. The dataset consists of synthetically-generated reasoning examples in natural language. \rev{Figure~\ref{fig:pipeline} illustrates the pipeline for generating each question.}
 We evaluate LLM's inductive and abductive reasoning capabilities on metrics such as accuracy and their adherence to Occam's Razor. By composing inductive and abductive reasoning steps shown in Table \ref{t:hypotheses}, our evaluation framework can readily 
 be extended to produce any abductive/inductive reasoning question expressible in first-order logic.

\begin{figure*}
\centering
\resizebox{\linewidth}{!}{
\begin{tikzpicture}[
node distance=2cm,
font=\normalsize,
box/.style={
    rectangle,
    rounded corners=4pt,
    draw=blue!50!black,
    fill=blue!7,
    thick,
    minimum width=4.3cm,
    minimum height=1.4cm,
    text width=4cm,
    align=center
},
bigbox/.style={
    rectangle,
    rounded corners=5pt,
    draw=blue!60!black,
    fill=blue!4,
    thick,
    minimum width=11cm,
    minimum height=3.1cm
},
arrow/.style={
    ->,
    thick,
    draw=black!70,
    -{Latex[length=3mm]}
}
]

\node[box] (s1)
{\textbf{Step 1 (\S\ref{sec:world_model})}\\ Generate the world model};

\node[box, right=of s1] (s2)
{\textbf{Step 2 (\S\ref{sec:obs_gen})}\\ Generate observations};

\node[bigbox, right=of s2] (s3) {};

\node[below=0.2cm of s3.north]
{\textbf{Step 3: Naturalistic Question Generation (\S\ref{sec:trans})}};

\node[box, anchor=west] (s3a)
at ([xshift=0.6cm,yshift=-0.3cm]s3.west)
{Convert theories and observations into natural language using templates};

\node[box, right=1.2cm of s3a] (s3b)
{Use \textsc{GPT-4o} to paraphrase examples into more naturalistic language};

\draw[arrow] (s1) -- (s2);
\draw[arrow] (s2) -- (s3.west);
\draw[arrow] (s3a) -- (s3b);

\end{tikzpicture}
}
\caption{\rev{Pipeline for generating each reasoning question in \dataset.}
}
\label{fig:pipeline}
\end{figure*}

Each reasoning example contains a fictional world model and a set of observations. We use real person names (such as \bluett{Amy}) and real property names (such as \bluett{strong}), but we do not use real concept names (such as \bluett{tiger}). Instead, we use fictional concept names (such as \bluett{wumpus}). In doing so, LLMs will not be confused by these fictional relations, as we never generate facts that contradict those about real concepts.

LLMs must produce hypotheses to \emph{explain} the observations. Our \emph{hypothesize-explain} framework can cover a wide range of inductive/abductive reasoning scenarios.\footnote{Traditionally, producing hypotheses to explain observations is considered abductive reasoning. However, we include inductive reasoning in the same framing. For example, given these axioms as a world model: \bluett{Sally is a cat;} \bluett{Polly is a cat;} \bluett{Amy is a cat} and these observations: \bluett{Sally is cute;} \bluett{Polly is cute;} \bluett{Amy is cute}, we can produce a hypothesis: \bluett{{All cats are cute}} to explain all observations. This is \emph{inductive reasoning}.}
To generate a single reasoning example, we go through the following steps:
\begin{enumerate}
    \item First, we construct a world model as an ontology tree in first-order logic.
    \item Second, we traverse the ontology tree and randomly mark some axioms as hidden. Each hidden axiom is the ground truth.
    \item Third, for each hidden axiom, we generate observations so that LLMs have enough information to produce ground-truth hypotheses.
    \item Fourth, we transform the incomplete world model and observations into natural language.
\end{enumerate}

\begin{table*}
\centering
    \footnotesize
    \scalebox{0.92}{
    \hspace{-0.9em}
    \begin{tabular}
    { 
    >{\raggedright\arraybackslash}m{0.1\textwidth} | 
    >{\raggedright\arraybackslash}m{0.11\textwidth} 
    >{\raggedright\arraybackslash}m{0.18\textwidth} 
    >{\raggedright\arraybackslash}m{0.12\textwidth} 
    >{\raggedright\arraybackslash}m{0.19\textwidth} 
    >{\centering\arraybackslash}m{0.24\textwidth} 
    } 
    \toprule[0.1em]
		\textbf{Subtask} & \textbf{Reasoning type tested} & \textbf{World model} &\textbf{Observations} &\textbf{Ground-truth hypothesis} &\textbf{Proof trees for observations} \\ 
    \midrule[0.1em]
    Infer property & Inductive & \texttt{o(A)} \newline \texttt{o(B)} \newline \texttt{o(C)} & \texttt{p(A)} \newline \texttt{p(B)} \newline \texttt{p(C)} & \(\forall x (o(x)\to p(x))\) & 
\vspace{.8em}
    
    $\prftree[r]{}{
                 \color{Purple}o(A)
              }{
                  \color{OliveGreen} \forall x(o(x) \to p(x))
              }{\color{black} p(A)}$

    \\
    \midrule
        Infer \newline membership \newline relation & Abductive & \(\forall x (o(x)\to p_1(x))\) \newline \(\forall x (o(x)\to p_2(x))\)  \newline  \(\forall x (o(x)\to p_3(x))\)  & \texttt{p1(A)} \newline \texttt{p2(A)} \newline \texttt{p3(A)} & \texttt{o(A)} & 
\vspace{.8em}
    
    $\prftree[r]{}{
                 \color{OliveGreen}o(A)
              }{
                  \color{Purple} \forall x(o(x) \to p_1(x))
              }{\color{black} p_1(A)}$

    \\

     \midrule
        Infer \newline subtype \newline relation & Both & \texttt{o1(A)} \newline \texttt{o1(B)}  \newline \texttt{o1(C)}  & \texttt{o2(A)} \newline \texttt{o2(B)} \newline \texttt{o2(C)} & \(\forall x(o_1(x)\to o_2(x))\) & 
\vspace{.8em}
    
    $\prftree[r]{}{
                 \color{Purple}o_1(A)
              }{
                  \color{OliveGreen} \forall x(o_1(x) \to o_2(x))
              }{\color{black} o_2(A)}$

    \\

    \bottomrule[0.1em]
	\end{tabular}
    }
    \caption{An overview of various subtasks and their details in the simplest case (a height-1 ontology tree with a single hypothesis). Here \texttt{o} denotes an ontology concept (e.g, \bluett{cat}) and \texttt{o(A)} denotes the membership (e.g., \bluett{cat(Sally)} means \bluett{Sally is a cat}). \texttt{p} denotes a property (e.g, \bluett{cute}) and \texttt{p(A)} denotes \texttt{A} has property \texttt{p} (e.g., \bluett{cute(Sally)} means \bluett{Sally is cute}). We only show the proof tree for one observation for simplicity; other observations can be derived similarly. In the proof tree, we show hypotheses in {\color{OliveGreen}{green}}  and axioms in world models in {\color{Purple}{purple}}.
    }
    \label{t:hypotheses}
\end{table*}

See Table \ref{t:hypotheses} for an overview of the different subtasks and their details in the simplest case (a height-1 ontology tree with a single hypothesis).
We evaluate the output of LLMs (i.e., produced \emph{hypotheses}) on the following metrics: \begin{enumerate}
    \item \emph{Strong accuracy}: Hypotheses are strongly correct if they exactly match ground-truth hypotheses (simplest hypothesis).
    \item \emph{Weak accuracy}: Hypotheses are weakly correct if they explain all observations.
    \item \emph{Hypothesis quality}: A quantitative measurement of the quality of the LLMs' produced hypotheses based on Occam's Razor.
\end{enumerate}

Next, we describe each step in more detail.

\subsection{World model generation}
\label{sec:world_model}
In our approach, the world model is an ontology tree in first-order logic. Each node in the ontology tree is a concept (e.g. \bluett{cat}) with properties (e.g. \bluett{All cats are cute}
) and members (e.g. \bluett{Amy is a cat}).
Each edge in the ontology tree denotes a subtype relationship (e.g., the \bluett{ragdoll} node being a child of the \bluett{cat} node implies \bluett{Each ragdoll is a cat}).

We first generate the topology of the ontology tree with the height as a controllable parameter. Each non-leaf node has a set of child nodes. The height of the ontology tree is a variable that characterizes the complexity of the world model. We then make two passes over the tree using level-order traversal. In the first pass, for each node, we allocate a concept name, properties, and members.\footnote{{The properties can also be negated, e.g., \bluett{All cats are not slow}, which corresponds to the logical form \(\forall x(\bluett{cat}(x) \rightarrow \neg\bluett{slow}(x))\) (negation always takes narrow scope)}.  {The number of properties, members, and child nodes follows a zipf distribution with \(a=3\).}}
After the first pass, we have a \emph{complete} world model with three types of axioms: 
\begin{enumerate*}[label=(\arabic*)]
\item properties (e.g., \bluett{All cats are cute}),
\item membership relations (e.g., \bluett{Amy is a cat}),
\item subtype relations (e.g., \bluett{Each ragdoll is a cat}).
\end{enumerate*}
In the second pass, we randomly mark some axioms as hidden. These axioms are the ground-truth hypotheses. After the second pass, we have an \emph{incomplete} world model.

\subsection{Observation generation}
\label{sec:obs_gen}
Next, we generate observations so that LLMs have sufficient information to produce the ground-truth hypotheses. The observations must be complex enough that the reasoning example is non-trivial, yet they should also ensure the problem is feasible.
\begin{itemize}
    \item To generate an example where the task is to infer a property (e.g., \bluett{All cats are cute}), we first identify the type in the ontology with this property (e.g., \bluett{cat}), then we sample members of this type and subtypes (e.g., \bluett{Amy is a cat}; \bluett{Sally is a ragdoll}; \bluett{Polly is a shorthair}), and add observations where the members have this property (e.g., \bluett{Amy is cute}; \bluett{Sally is cute};).
    \item To generate an example where the task is to infer a membership relation (e.g., \bluett{Amy is a cat}), we first identify the type in the ontology with this member (e.g., \bluett{cat}), then we sample properties of this type and supertypes (e.g., \bluett{All cats are cute}; \bluett{All felines are not slow}; \bluett{All mammals are hairy}), and add observations where this member have these properties (e.g., \bluett{Amy is cute}; \bluett{Amy is not slow}).
    \item  To generate an example where the task is to infer a subtype relation (e.g., \bluett{All cats are felines}), we first identify the subtype in the ontology (e.g., \bluett{cat}), then we sample members of this type and its subtypes (e.g., \bluett{Amy is a cat}; \bluett{Sally is a ragdoll}; \bluett{Polly is a shorthair}), and add observations where these members belong to the supertype (e.g., \bluett{Amy is a feline}; \bluett{Polly is a feline}).
\end{itemize}

\subsection{Naturalistic Question Generation}
\label{sec:trans}
We first use the same grammar as in \textsc{PrOntoQA} for converting logical forms into natural language using templates. Properties are converted into the form: 
 \bluett{All cats are cute}; subtype relations are randomly converted into one of three forms: \bluett{Each ragdoll is a cat;} \bluett{Every ragdoll is a cat;} \bluett{All ragdolls are cats}; membership relations are converted into the form \bluett{Amy is a cat}. We then use \texttt{GPT-4o} to paraphrase into more natural language. We also manually verify that the generated questions from \textsc{GPT-4o} are sound and do not alter the problem's semantics. We conduct this study and iteratively improve the prompt until achieving 100\% accuracy. \rev{Details are provided in Appendix~\ref{sec:translate}.}

\subsection{Hypothesis quality}
We define a metric to evaluate the quality of hypotheses 
 based on the principles of Occam's Razor (i.e., the priciple that, other things being equal, simpler theories are better; ~\citealt{occ}).   
 In this regard, we encourage one hypothesis to explain as many observations as possible and penalize unnecessary hypotheses. 
 Formally, given a world model \(\mathcal T\), a set of observations \(\mathcal O\), their associated proof trees \(\mathcal P\),candidate hypotheses \(\mathcal H\), and ground-truth hypotheses \(\mathcal H^*\), for \(h \in \mathcal H\), \(n(h)\) denotes the number of times \(h\) appears in the proof trees of \(\mathcal O\) (i.e., how many times \(h\) is used as a premise of a proof step). We first use \(|\mathcal H|\) to denote the cardinality of \(\mathcal H\). The quality of \(\mathcal H\), \(q(\mathcal H)\) is defined by \footnote{We write the equation as the ratio of two averages: the numerator is the average accuracy of the predicted hypothesis (i.e., the average number of times a hypothesis appears in any valid proof), and the denominator is the average number of times a hypothesis appears in the shortest (i.e., most parsimonious) proof.}:
 \[
 q(\mathcal H) = \begin{cases}
     \frac{\frac{1}{|\mathcal H|}\sum_{h \in \mathcal H} n(h)}
     {\frac{1}{|\mathcal H^*|}\sum_{h^* \in \mathcal H^{*}} n(h^*)} & \mathcal{H} \ \text{can explain} \ \mathcal{O} \\
     0 & \text{otherwise} \\
 \end{cases}
 \]

According to the definition, the ground-truth hypotheses \(\mathcal H^*\) have a quality score of 1. Consider the example in Figure \ref{fig:banner}, here \(\mathcal H^*\) is \{\bluett{Fae is a tiger}, \bluett{All mammals are hairy}, \bluett{All rodents are mammals}\} and \(|\mathcal H^*| = 3\). Each \(h^* \in \mathcal H^*\) appears three times in the proof trees of \(\mathcal O\), so \(\sum_{h^* \in \mathcal H^*}n(h^*) = 9\). Consider candidate hypotheses \(\mathcal H\) \{\bluett{Fae is a tiger}, \bluett{All mammals are hairy}, \bluett{All rats are mammals}, \bluett{All squirrels are mammals}, \bluett{All cats are hairy}\}. \(|\mathcal H| = 5\). Obviously \(\mathcal H\) can explain \(\mathcal O\). \bluett{All rats are mammals} appears twice, \bluett{All squirrels are mammals} appears once, and \bluett{All cats are hairy} appear once in the proof trees of \(\mathcal O\), so \(\sum_{h \in \mathcal H}n(h) = 3 + 3 + 2 + 1 + 1 = 10\). In this case, \(q(\mathcal H) = \frac{3 \times 10}{5 \times 9} = 0.67\).
 
 \rev{We further motivate the hypothesis metric using Bayesian decision theory in Appendix~\ref{sec:metric_theory}. We compare the metric against human evaluation of the hypothesis; the results and annotator details are in Appendix~\ref{sec:metric_human_eval}.}  
We also include the annotator instructions in Figure~\ref{fig:instruction_annotator}. Appendix~\ref{sec:algo} contains the complete algorithms for generating reasoning examples and evaluating hypotheses.

\section{ \old{Results} \rev{Experiments}}
\subsection{Experimental setup}
In each experiment configuration, we generate 100 independent and identically distributed reasoning examples. Each reasoning example is pure text, a user prompt with an incomplete world model and observations. We also specify the desired output format in the system prompt. To answer each reasoning example, LLMs must produce hypotheses to explain all observations. We evaluate hypotheses on three metrics: weak accuracy, strong accuracy, and quality. We compute 95\%  confidence intervals for both accuracy metrics using the Wilson interval~\citep{wilson}.   
We use the OpenAI API and \href{https://www.together.ai/}{together.ai} platform.

In our experiments, we vary the following variables: \begin{enumerate*}[label=(\arabic*)]
    \item Different types of tasks (i.e., infer property, infer membership relation, or infer subtype relation),
    \item Single hypothesis or multiple hypotheses,
    \item The height of the ontology tree,
    \item Whether the reasoning example contains in-context demonstrations and whether they are in-distribution,
    \item \rev{Various reasoning-augmented prompts such as buffer-of-thought~\citep{bot}}.
\end{enumerate*}

\subsection{Zero-shot results}
We first investigate the simplest setting with zero-shot prompting.
\subsubsection{Single hypothesis}
We first investigate results for reasoning examples with one hypothesis for three tasks. This helps us to isolate different types of tasks and understand how LLMs perform in each case. The results are in Figure \ref{fig:single}. In general, all models show high accuracy (both weak and strong) greater than 80\% 
for all three tasks under a height-1 ontology tree. When the height of the ontology tree starts to increase, we also observe a general decreasing trend in all three metrics, which is not surprising. One interesting observation is that LLMs perform better in \emph{infer membership relation} tasks than other tasks, especially for \textsc{Gemma3-27B}. For example, even under a height-4 ontology tree, \textsc{Gemma3-27B} can achieve near 50\% weak accuracy, while the number drops to around 10\% for \emph{infer subtype relation} tasks. This may be because the weak correctness for \emph{infer membership relation} tasks only requires searching for the concept that has some property, but do not need to understand the structure of the ontology tree.\footnote{e.g., given \bluett{Amy is cute}, LLMs only need to search for a concept which has the property \bluett{cute} to infer the membership relation of \bluett{Amy}.} However, this is not the case for \emph{infer subtype relation} and \emph{infer properties} tasks. We observe a similar trend for \emph{infer subtype relation} and \emph{infer properties} tasks, as we can view the concept's parent as a special property.

\begin{figure*}
    \centering
    \resizebox{\textwidth}{!}{\input{figures/single.pgf}}
    \caption{Weak accuracy, strong accuracy, and quality versus the height of the ontology tree with single hypothesis.}
    \label{fig:single}
\end{figure*}

\begin{figure*}
    \centering
    \resizebox{\textwidth}{!}{\input{figures/icl.pgf}}
        \caption{Weak accuracy, strong accuracy, and quality versus the height of the ontology tree with multiple hypotheses under zero-shot (\textbf{top}), 8-shot in-distribution (\textbf{middle}), and 8-shot out-of-distribution (\textbf{bottom}). 
        We list the absolute results of the 8-shot in-distribution and out-of-distribution in Appendix~\ref{sec:moreresults}.
        }
    \label{fig:icl}
\end{figure*}

\subsubsection{Multiple hypotheses}
Next, we turn to multi-hypothesis scenarios because they are closer to real-life situations, and we are interested in understanding how LLMs perform in this setting. Each reasoning example can contain multiple hypotheses. To check the difficulty of each reasoning example, we sample the ground-truth hypotheses such that the number of hypotheses grows linearly \emph{w.r.t.} the height of the ontology tree.\footnote{Each ground-truth hypothesis has $\sim3$ observations. So the number of observations also grows linearly \emph{w.r.t.} the height of the ontology tree.} Invariably, the number of axioms in the world model grows exponentially \emph{w.r.t.} the height of the ontology tree. See Table \ref{t:statistics} in Appendix~\ref{sec:basic_statistics} for statistics under multi-hypothesis scenarios and the \textbf{top} row of Figure \ref{fig:icl} for results.

Performance in the multi-hypothesis setting differed significantly from that in the single-hypothesis setting. For example, the accuracy of all models drops significantly when the height of the ontology tree is increased from 1 to 2, with accuracy going from above 80\% to below 50\%, except for \textsc{GPT-4o}. 
This is surprising because the average number of ground-truth hypotheses increases only from \(3.0\) to \(3.5\).  We attribute this to the coupling effect of multiple hypotheses to produce and the topology of the ontology tree. Reasoning examples under the ontology tree with heights 3 or 4 achieve lower accuracy and quality scores across all models.  We also observe a notable gap between weak accuracy and quality. For example, under the height-4 ontology tree, all models can achieve at least 20\% weak accuracy but struggle with extremely low-quality and strong-accuracy scores. We manually inspect some incorrect hypotheses and summarize the most common types of errors in Section \ref{sec:err}.

\subsection{Does in-context learning help?}

LLMs have shown extraordinary generalization capabilities with few-shot examples coupled with CoT prompts~\citep{iclclassicla}. We are also interested in whether in-context learning can help with inductive/abductive reasoning.  We add $8$ demonstrations to each example, with the demonstrations' ground-truth hypotheses and CoTs. The CoT consists of each observation's step-by-step proof. We have two settings for in-context learning: 
\begin{enumerate*}[label=(\arabic*)]
    \item In-distribution demonstrations, where the demonstration has an ontology tree of the same height as the test question.
    \item Out-of-distribution demonstrations, where each demonstration has a single hypothesis with a height-1 ontology tree. 
\end{enumerate*} 
In both settings, we ensure demonstrations cover all types of tasks. \rev{We present our results in Figure~\ref{fig:icl}.}

The in-distribution demonstrations are slightly more helpful than out-of-distribution demonstrations. This result is consistent with the hypothesis that in-context learning mimics Bayesian inference~\citep{cotbay}. We also observe that in-context learning with in-distribution demonstrations improves strong accuracy and quality in ontology trees with heights 3 and 4. We conjecture that LLMs can learn how to navigate the ontology hierarchy and produce high-quality hypotheses through in-context demonstrations. In contrast, we do not observe a significant performance improvement from out-of-distribution demonstrations.

\subsection{\rev{Do reasoning models and reasoning-augmented prompts help?}}
\begin{figure*}[ht]
    \centering
    \resizebox{\textwidth}{!}{
    \includegraphics[width=\textwidth]{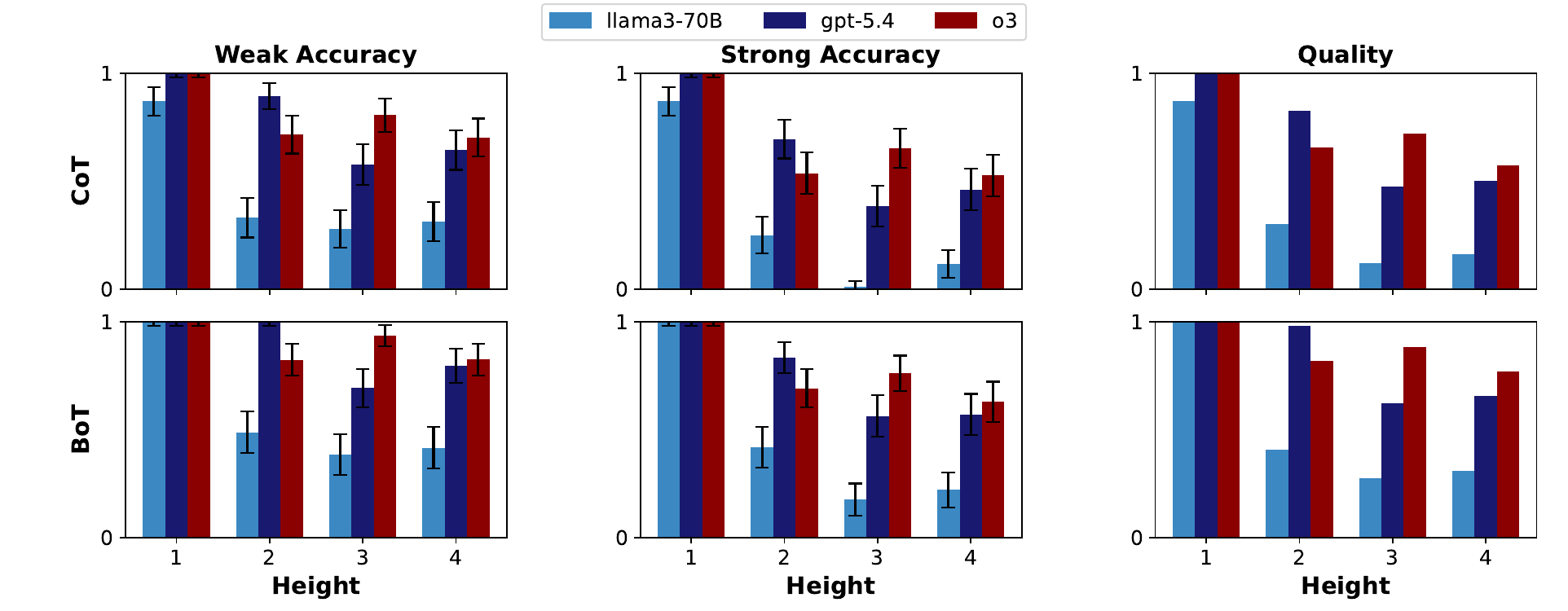}}
    \caption{Results for \textsc{Llama3-70B} and two large reasoning models \textsc{gpt-5.4} and \textsc{o3} with chain-of-thought prompting (\textbf{top}) and buffer-of-thought prompting (\textbf{bottom}).}
    \label{fig:rlvr}
\end{figure*}
\rev{Recent developments in large reasoning models, particularly those trained with RLVR, have led to substantial improvements across many evaluations. As shown in Figure~\ref{fig:rlvr}, stronger models such as \textsc{gpt5.4} and \textsc{o3} consistently outperform \textsc{LLaMA3-70B} in both weak and strong accuracy and overall quality across different ontology tree heights. Moreover, reasoning-oriented prompting strategies further enhance performance: buffer-of-thought~\citep[BoT;][]{bot} prompting generally yields higher scores than chain-of-thought prompting~\citep{CoT}, especially in weak accuracy.\footnote{Appendix~\ref{sec:reasoning_augmented_prompts} lists the reasoning-augmented prompts used in our experiments.} However, despite these gains, a persistent gap remains between weak and strong accuracy across all models and settings—--for instance, weak accuracy often approaches near-perfect levels at lower heights, while strong accuracy remains notably lower and degrades more rapidly as height increases. This suggests that although recent advances and prompting techniques significantly improve reasoning capabilities, high-quality hypothesis generation remains an open challenge.}

\subsection{Analysis of common errors}
\label{sec:err}
To further understand \emph{how} and \emph{why} LLMs fail, we manually inspect mistaken and low-quality hypotheses. We summarize some of the most common types of errors here for further insights. We provide example responses in each category, quantitative analysis, and the details of the manual inspection process in Appendix \ref{sec:response}.
\begin{enumerate*}[label=(\arabic*)]
\item \textbf{Wrong ontology direction.}
 LLMs sometimes produce hypotheses in the wrong direction. For example, the ground-truth hypothesis is \bluett{All cats are mammals}, while LLMs will produce the hypothesis \bluett{All mammals are cats}.
\item \textbf{Ignore the ontology and produce unnecessary hypotheses.} LLMs sometimes can produce unnecessary hypotheses because they ignore the ontology. 
\item \textbf{Fall back to trivial hypotheses.} Technically, the observation is also a valid hypothesis, but it is considered low-quality because it can only explain itself. LLMs sometimes reuse observations as hypotheses.
\item \textbf{Hallucinated entities.} LLMs sometimes produce hypotheses with hallucinated entities (including concepts, properties, members). Usually, this happens when the entity appears in the in-context demonstrations, but LLMs incorrectly re-use them in the test question.
\end{enumerate*}
\subsection{Improving non-monotonic reasoning via synthetic data post-training}

\begin{figure}[ht]
    \centering
    \includegraphics[width=\columnwidth]{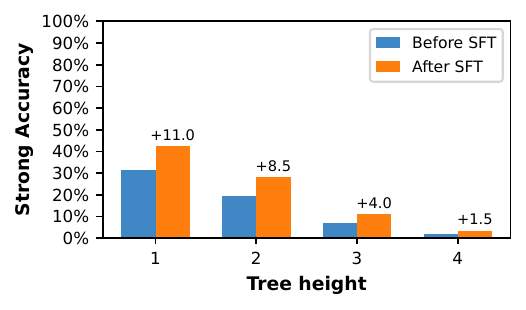}
    \caption{Strong accuracy before and after supervised fine-tuning (SFT)
    across ontology-tree heights.}
    \label{fig:sft_results}
\end{figure}

We show that our synthetic data can improve inductive and abductive reasoning of LLMs by performing supervised fine-tuning on \textsc{Qwen3-4B}~\citep{qwen3} on the training set and evaluating it on a held-out test set. We use \texttt{Unsloth}~\citep{unsloth} and \texttt{TRL}~\citep{trl} for training. Figure~\ref{fig:sft_results} shows improvements across all tree heights, ranging from \(+1.5\%\) at height~4 to \(+11.0\%\) at height~1. Table~\ref{tab:sft_hyperparameters} lists the training hyperparameters.

\begin{table}[t]
\centering
\small
\begin{tabular}{lc}
\toprule
\textbf{Hyperparameter} & \textbf{Value} \\
\midrule
Learning rate & \(2 \times 10^{-4}\) \\
Optimizer & AdamW \\
LoRA rank & 16 \\
Training epochs & 1 \\
Weight decay & 0.01 \\
\bottomrule
\end{tabular}
\caption{Hyperparameters used for SFT.}
\label{tab:sft_hyperparameters}
\end{table}
\section{Conclusion}
In this study, we systematically examined the inductive/abductive reasoning capabilities of LLMs, especially whether they follow Occam's Razor.  Our results suggest that current LLMs can’t generate high-quality hypotheses that match human-level parsimony. Our results can serve to better inform practitioners of LLMs of their limitations in inductive/abductive reasoning. The proposed dataset opens new avenues for future work, such as further post-training models to improve their non-monotonic reasoning capabilities, for example via reinforcement learning.

\section*{Limitations}
 Our synthetic dataset serves as an out-of-distribution evaluation of the inductive and abductive reasoning abilities of LLMs. We do not claim that success in our benchmark automatically implies success in real-world tasks. Still, our benchmark serves as a checkpoint for developing models that are capable of robust real-world abductive and inductive reasoning.

Another limitation of this work is that first-order logic (FOL) restricts the expressiveness of the world model and the reasoning example. We leave it as future work to utilize higher-order logic (HOL) to generate more complex world models and reasoning examples.

\old{In this work, we analyze common errors by hand from randomly selected examples. Although this approach provides many essential insights into where LLMs fail, it is not scalable. We leave it as future work to leverage more scalable methods, such as LLM-as-a-Judge, to analyze common errors at a larger scale.}

\bibliography{custom}
\appendix

\raggedbottom

\section{Preliminary background}
\label{sec:prelim}
\subsection{Examples of different types of reasoning}

\rev{For example, given ``All humans are moral'' and ``Socrates is a human'', we can conclude that ``Socrates is moral''.  For example, if we see many cute cats, we may inductively reason that all cats are cute.  For example, if we hear a bird sound outside, we may abductively reason that there is a bird. The latter two types of reasoning share a common high-level goal: to find explanations for observations. Furthermore, the composition of inductive and abductive reasoning is prevalent in real-world scenarios. For example, given (i) {Amy has a skin rash and leg pain and has been diagnosed with a new, unknown virus.} (ii) {Sam has a skin rash and leg pain and has been diagnosed with the same new unknown virus.} (iii) {Tom also has a skin rash and leg pain.} We can first induce the rule: {The new unknown virus causes skin rash and leg pain}, and then we can abduce that {Tom is infected with this new virus}.}
\subsection{Proof trees}
\label{sec:appendix_prooftree}

\rev{A \textbf{proof tree} is a tree-structured representation of a logical derivation where each node corresponds to a statement, and each edge represents the application of an inference rule. The root of the tree is the final conclusion, while the leaves correspond to premises or axioms. Proof trees are commonly used in logic, theorem proving, and formal reasoning to make the structure of derivations explicit.}

\rev{Formally, let $\Gamma$ denote a set of premises and $\varphi$ denote a conclusion. A proof tree demonstrates that $\Gamma \vdash \varphi$ by recursively applying inference rules until all leaf nodes correspond to elements of $\Gamma$ or axioms.}

\paragraph{Example.}
\rev{Consider the following premises:}

\begin{align*}
1.\quad & P \rightarrow Q \\
2.\quad & Q \rightarrow R \\
3.\quad & P
\end{align*}

\rev{We can derive $R$ using two applications of \emph{modus ponens}. We first derive $Q$ given the premises $P$ and $P \rightarrow Q$. Next, we derive $R$ using the premises $Q$ and $Q \rightarrow R$. This proof can be represented as a proof tree, which is shown below.}

\[
\frac{
    \frac{P \rightarrow Q \qquad P}{Q}
    \qquad
    Q \rightarrow R
}{
    R
}
\]

\rev{The tree therefore makes explicit the hierarchical structure of the reasoning process.}
\section{Basic statistics of \dataset}
We list basic statistics of \dataset \  in Table~\ref{t:statistics}.
\label{sec:basic_statistics}
\begin{table}[t]
\centering
    \footnotesize    
    \makebox[\linewidth][c]{
	\def\arraystretch{0.9}
    \begin{tabular}
    { 
    >{\raggedright\arraybackslash}m{0.13\textwidth}  
    >{\raggedright\arraybackslash}m{0.13\textwidth} 
    >{\raggedright\arraybackslash}m{0.14\textwidth} 
    >{\raggedright\arraybackslash}m{0.20\textwidth} 
     >{\raggedright\arraybackslash}m{0.12\textwidth}
    } 
    \toprule[0.1em]
		\textbf{Ontology tree height} & \textbf{\# of world model axioms} & \textbf{\# of observations} &\textbf{\# of ground-truth hypotheses} & \textbf{\# of examples}\\ 
    \midrule[0.1em]
    1 & 9.0 & 10.0 & 3.0 & 500 \\
   \midrule
    2 & 14.0 & 11.8 & 3.5 & 500\\
   \midrule
    3 & 25.5 & 15.2 & 4.6 & 500 \\
   \midrule
    4 & 46.8 & 20.0 & 6.6 & 500 \\
    \bottomrule[0.1em]
	\end{tabular}
    }
    \caption{Basic statistics of multi-hypothesis examples in \dataset. We list the average number of world model axioms, observations, and ground-truth hypotheses \emph{w.r.t.} the height of the ontology tree in the same set of reasoning examples we used in evaluation. Note that the number of the ground-truth hypotheses and the number of observations grow linearly with the height of the ontology tree. In contrast, the number of world model axioms grows exponentially.
    }
    \label{t:statistics}
\end{table}
\section{More examples of reasoning questions}
\label{sec:more_examples}

In this section, we provide additional reasoning examples generated using our proposed framework. For each setting listed below, we provide an example consisting of a world model, list of observations, and ground-truth hypothesis/hypotheses.


\begin{tcolorbox}[
colback=gray!5,
colframe=black!60,
arc=3pt,
boxrule=0.6pt,
breakable,
left=4pt,right=4pt,
title=\textbf{\emph{Single hypothesis example; task: infer property; height-1 ontology tree.}}
]
            \textbf{World Model:} {Amy is a dalpist. Pamela is a dalpist. Jerry is a dalpist.}
            \\
            \textbf{Observations:} {Amy is rainy. Jerry is rainy. Pamela is rainy.}
            \\
            \textbf{Hypotheses:} {Dalpists are rainy.}
\end{tcolorbox}

\begin{tcolorbox}[
colback=gray!5,
colframe=black!60,
arc=3pt,
breakable,
boxrule=0.6pt,
left=4pt,right=4pt,
title=\textbf{\emph{Single hypothesis example; task: infer property; height-2 ontology tree.}}
]
            \textbf{World Model:} {Rompuses are dalpists. Sarpers are dalpists. Each gergit is a dalpist. Pamela is a sarper. Jerry is a dalpist. Amy is a rompus.}
            \\
            \textbf{Observations:} {Amy is rainy. Pamela is rainy. Jerry is rainy.}
            \\
            \textbf{Hypotheses:} {Dalpists are rainy.}
\end{tcolorbox}

\begin{tcolorbox}[
colback=gray!5,
colframe=black!60,
arc=3pt,
breakable,
boxrule=0.6pt,
left=4pt,right=4pt,
title=\textbf{\emph{Single hypothesis example; task: infer property; height-3 ontology tree.}}
]
            \textbf{World Model:} {Pamela is a porpor. Jerry is a dalpist. Twimpees are sarpers. Each gergit is a dalpist. Each storpist is a gergit. Amy is a storpist. Porpors are rompuses. Sarpers are dalpists. Rompuses are dalpists.
}
            \\
            \textbf{Observations:} {Jerry is not muffled. Amy is not muffled. Pamela is not muffled.}
            \\
            \textbf{Hypotheses:} {Each dalpist is not muffled.}
\end{tcolorbox}

\begin{tcolorbox}[
colback=gray!5,
colframe=black!60,
breakable,
arc=3pt,
boxrule=0.6pt,
left=4pt,right=4pt,
title=\textbf{\emph{Single hypothesis example; task: infer property; height-4 ontology tree.}}
]
            \textbf{World Model:} {Amy is a felper. Jerry is a dalpist. Pamela is a felper. Each rompus is a dalpist. Each porpor is a rompus. Sarpers are dalpists. Shumples are twimpees. Gergits are dalpists. Each frompor is a porpor. Storpists are gergits. Twimpees are sarpers. Felpers are storpists.
}
            \\
            \textbf{Observations:} {Amy is translucent. Pamela is translucent. Jerry is translucent.
}
            \\
            \textbf{Hypotheses:} {Each dalpist is translucent.}
\end{tcolorbox}

\begin{tcolorbox}[
colback=gray!5,
colframe=black!60,
arc=3pt,
breakable,
boxrule=0.6pt,
breakable,
left=4pt,right=4pt,
title=\textbf{\emph{Single hypothesis example; task: infer membership relation; height-1 ontology tree.}}
]
\textbf{World Model:} Dalpists are liquid. Dalpists are rainy. Each dalpist is brown.
\\
\textbf{Observations:} Jerry is brown. Jerry is rainy. Jerry is liquid.
\\
\textbf{Hypotheses:} Jerry is a dalpist.
\end{tcolorbox}

\begin{tcolorbox}[
colback=gray!5,
colframe=black!60,
arc=3pt,
breakable,
boxrule=0.6pt,
left=4pt,right=4pt,
title=\textbf{\emph{Single hypothesis example; task: infer membership relation; height-2 ontology tree.}}
]
\textbf{World Model:} Rompuses are dalpists. Jerry is a dalpist. Each sarper is a dalpist. Each dalpist is brown. Edward is a sarper. Gergits are dalpists. Gergits are rainy. Amy is a rompus. Dalpists are liquid.
\\
\textbf{Observations:} Pamela is rainy. Pamela is liquid. Pamela is brown.
\\
\textbf{Hypotheses:} Pamela is a gergit.
\end{tcolorbox}

\begin{tcolorbox}[
colback=gray!5,
colframe=black!60,
arc=3pt,
boxrule=0.6pt,
breakable,
left=4pt,right=4pt,
title=\textbf{\emph{Single hypothesis example; task: infer membership relation; height-3 ontology tree.}}
]
\textbf{World Model:} Each sarper is a dalpist. Jerry is a dalpist. Barbara is a porpor. Each dalpist is moderate. Rompuses are dalpists. Twimpees are sarpers. Storpists are gergits. Each dalpist is large. Edward is a sarper. Pamela is a gergit. Gergits are dalpists. Debra is a storpist. Porpors are rompuses. Amy is a rompus. Twimpees are not muffled.
\\
\textbf{Observations:} Nicole is moderate. Nicole is large. Nicole is not muffled.
\\
\textbf{Hypotheses:} Nicole is a twimpee.
\end{tcolorbox}

\begin{tcolorbox}[
colback=gray!5,
colframe=black!60,
arc=3pt,
breakable,
boxrule=0.6pt,
left=4pt,right=4pt,
title=\textbf{\emph{Single hypothesis example; task: infer membership relation; height-4 ontology tree.}}
]
\textbf{World Model:} Amy is a rompus. Each porpor is a rompus. Each sarper is a dalpist. Sharon is a felper. Rompuses are dalpists. Each shumple is a twimpee. Each dalpist is fruity. Barbara is a porpor. Every frompor is a porpor. Every gergit is a dalpist. Edward is a sarper. Raymond is a frompor. Each storpist is a gergit. Jerry is a dalpist. Each dalpist is moderate. Shumples are translucent. Felpers are storpists. Nicole is a twimpee. Pamela is a gergit. Twimpees are sarpers. Debra is a storpist.
\\
\textbf{Observations:} Michael is fruity. Michael is translucent. Michael is moderate.
\\
\textbf{Hypotheses:} Michael is a shumple.
\end{tcolorbox}

\begin{tcolorbox}[
colback=gray!5,
colframe=black!60,
arc=3pt,
breakable,
boxrule=0.6pt,
left=4pt,right=4pt,
title=\textbf{\emph{Single hypothesis example; task: infer subtype relation; height-1 ontology tree.}}
]
\textbf{World Model:} Amy is a dalpist. Pamela is a dalpist. Jerry is a dalpist.
\\
\textbf{Observations:} Amy is a rompus. Jerry is a rompus. Pamela is a rompus.
\\
\textbf{Hypotheses:} Dalpists are rompuses.
\end{tcolorbox}

\begin{tcolorbox}[
colback=gray!5,
colframe=black!60,
arc=3pt,
boxrule=0.6pt,
breakable,
left=4pt,right=4pt,
title=\textbf{\emph{Single hypothesis example; task: infer subtype relation; height-2 ontology tree.}}
]
\textbf{World Model:} Gergits are dalpists. Porpors are dalpists. Each sarper is a dalpist. Pamela is a porpor. Jerry is a dalpist. Amy is a gergit.
\\
\textbf{Observations:} Amy is a rompus. Pamela is a rompus. Jerry is a rompus.
\\
\textbf{Hypotheses:} Dalpists are rompuses.
\end{tcolorbox}

\begin{tcolorbox}[
colback=gray!5,
colframe=black!60,
arc=3pt,
boxrule=0.6pt,
breakable,
left=4pt,right=4pt,
title=\textbf{\emph{Single hypothesis example; task: infer subtype relation; height-3 ontology tree.}}
]
\textbf{World Model:} Amy is a storpist. Jerry is a dalpist. Frompors are porpors. Each sarper is a dalpist. Each twimpee is a sarper. Pamela is a twimpee. Storpists are gergits. Porpors are dalpists. Gergits are dalpists.
\\
\textbf{Observations:} Jerry is a rompus. Amy is a rompus. Pamela is a rompus.
\\
\textbf{Hypotheses:} Each dalpist is a rompus.
\end{tcolorbox}

\begin{tcolorbox}[
colback=gray!5,
colframe=black!60,
arc=3pt,
boxrule=0.6pt,
breakable,
left=4pt,right=4pt,
title=\textbf{\emph{Single hypothesis; task: infer subtype relation; height-4 ontology tree.}}
]
\textbf{World Model:} Each sarper is a dalpist. Jerry is a dalpist. Amy is a shumple. Pamela is a shumple. Each storpist is a gergit. Dumpuses are frompors. Gergits are dalpists. Felpers are storpists. Each twimpee is a sarper. Frompors are porpors. Shumples are twimpees. Porpors are dalpists.
\\
\textbf{Observations:} Amy is a rompus. Pamela is a rompus. Jerry is a rompus.
\\
\textbf{Hypotheses:} Each dalpist is a rompus.
\end{tcolorbox}

\begin{tcolorbox}[
colback=gray!5,
colframe=black!60,
arc=3pt,
breakable,
boxrule=0.6pt,
left=4pt,right=4pt,
title=\textbf{\emph{Multiple hypothesis example; height-1 ontology tree.}}
]
\textbf{World Model:} Debra is a dalpist. Amy is a dalpist. Pamela is a dalpist. Edward is a dalpist. Dalpists are liquid. Each dalpist is moderate. Each dalpist is brown. Barbara is a dalpist.
\\
\textbf{Observations:} Barbara is rainy. Jerry is a rompus. Jerry is liquid. Jerry is brown. Pamela is a rompus. Edward is rainy. Debra is rainy. Jerry is moderate. Amy is a rompus.
\\
\textbf{Hypotheses:} Dalpists are rompuses. Dalpists are rainy. Jerry is a dalpist.
\end{tcolorbox}

\begin{tcolorbox}[
colback=gray!5,
colframe=black!60,
arc=3pt,
breakable,
boxrule=0.6pt,
left=4pt,right=4pt,
title=\textbf{\emph{Multiple hypothesis example; height-2 ontology tree.}}
]
\textbf{World Model:} Nicole is a gergit. Porpors are dalpists. Pamela is a sarper. Sharon is a gergit. Porpors are not cold. Dalpists are rainy. Raymond is a gergit. Each porpor is translucent. Each sarper is a dalpist. Jerry is a dalpist. Edward is a dalpist. Every gergit is a dalpist. Debra is a dalpist. Barbara is a sarper.
\\
\textbf{Observations:} Barbara is liquid. Amy is translucent. Sharon is moderate. Debra is liquid. Amy is rainy. Raymond is moderate. Jerry is a rompus. Amy is not cold. Pamela is a rompus. Nicole is moderate. Edward is liquid. Amy is a rompus.
\\
\textbf{Hypotheses:} Dalpists are rompuses. Dalpists are liquid. Every gergit is moderate. Amy is a porpor.
\end{tcolorbox}

\begin{tcolorbox}[
colback=gray!5,
colframe=black!60,
breakable,
arc=3pt,
boxrule=0.6pt,
left=4pt,right=4pt,
title=\textbf{\emph{Multiple hypothesis example; height-3 ontology tree.}}
]
\textbf{World Model:} Barbara is a sarper. Sarpers are dalpists. Debra is a shumple. Every felper is a sarper. Raymond is a porpor. Sharon is a storpist. Each storpist is a gergit. Helen is a frompor. Porpors are dalpists. Each dalpist is liquid. Amy is a sarper. Each dalpist is salty. Felpers are rainy. Every shumple is a porpor. Each twimpee is a sarper. Nicole is a gergit. Each frompor is a sarper. Jerry is a dalpist. Dalpists are not muffled. Each gergit is a dalpist. Michael is a twimpee. Sarpers are amenable.
\\
\textbf{Observations:} Debra is large. Amy is a rompus. Pamela is a rompus. Edward is liquid. Pamela is rainy. Edward is not muffled. Barbara is large. Pamela is not muffled. Edward is large. Edward is salty. Jerry is a rompus. Pamela is amenable.
\\
\textbf{Hypotheses:} Each dalpist is a rompus. Dalpists are large. Edward is a dalpist. Pamela is a felper.
\end{tcolorbox}

\begin{tcolorbox}[
colback=gray!5,
colframe=black!60,
arc=3pt,
breakable,
boxrule=0.6pt,
left=4pt,right=4pt,
title=\textbf{\emph{Multiple hypothesis example; height-4 ontology tree.}}
]
\textbf{World Model:} Every kurpor is a twimpee. Each gergit is pale. Patricia is a dumpus. Timples are frompors. Sharon is a folpee. George is an orgit. Rimpees are felpers. Stephen is a shumple. Mark is an orgit. Joseph is a porpor. Andrew is an orgit. Each tumpus is a felper. Gergits are dalpists. Felpers are not dark. Helen is a timple. Jason is a tumpus. Amy is a pergit. Rimpees are angry. Porpors are dalpists. Raymond is a twimpee. Each rimpee is moderate. Linda is a tumpus. Every shumple is a porpor. Each frompor is a sarper. Every dumpus is a storpist. Gergits are amenable. Barbara is an orgit. Rachel is a frompor. Felpers are sarpers. Jacob is a felper. Janet is a tumpus. Michael is a timple. Steven is a boompist. Edward is a dalpist. Storpists are gergits. Each boompist is a frompor. Debra is a dalpist. Nicole is a sarper. Pergits are felpers. Shirley is a timple. Jerry is a dalpist. Pamela is a kurpor. Every folpee is a frompor. Each sarper is a dalpist. Eric is a storpist. Orgits are shumples. Gergits are not opaque. Twimpees are sarpers.
\\
\textbf{Observations:} Linda is brown. Helen is salty. George is translucent. Jason is brown. Raymond is large. Sharon is large. Jerry is a rompus. William is angry. Barbara is liquid. Edward is liquid. Emily is amenable. Janet is brown. Emily is not opaque. Shirley is salty. William is not dark. Pamela is a rompus. Nicole is large. William is moderate. Andrew is translucent. Michael is salty. Amy is a rompus. Debra is liquid. Emily is pale. Mark is translucent.
\\
\textbf{Hypotheses:} Dalpists are rompuses. Every dalpist is liquid. Emily is a gergit. Sarpers are large. Timples are salty. William is a rimpee. Tumpuses are brown. Orgits are translucent.
\end{tcolorbox}

\section{Details of \textsc{GPT-4o} paraphrase}
\label{sec:translate}
\subsection{Prompts}
\rev{We present our prompts for \textsc{GPT-4o} to paraphrase synthetic reasoning questions into more natural ones in Figure~\ref{fig:rewrite_prompt}.}
\begin{figure}[ht]

\begin{tcolorbox}[
colback=gray!5,
colframe=black!60,
arc=3pt,
boxrule=0.6pt,
title=\textbf{Prompt for Question Rewriting}
]

\textbf{Instruction}

We prompt \textsc{GPT-4o} to rewrite synthetic reasoning questions into more natural, human-readable language while preserving the original semantics and reasoning structure.


\textbf{Prompt Template}

\ttfamily
You are a professional editor specializing in rewriting synthetic reasoning questions into natural, human-readable language.

Rewrite the following question to be clear, fluent, and natural while preserving all logical structure and meaning. Do not simplify the reasoning steps, remove constraints, or change the problem's difficulty.

Return only the improved question with the exact same semantics, but in more natural language.

Input question: \{\{QUESTION\}\}

\normalfont
\end{tcolorbox}

\caption{Prompt used to paraphrase synthetic reasoning questions into more natural language.}

\label{fig:rewrite_prompt}

\end{figure}
\subsection{Manual inspection}
\rev{We also manually verify that the generated questions from \textsc{GPT-4o} are sound by randomly sampling 200 reasoning questions of varying difficulty and manually checking that \textsc{GPT-4o}'s output does not alter the problem's semantics. We conduct this study and iteratively improve the prompt until achieving 100\% accuracy.}

\subsection{Synthetic vs.\ paraphrased reasoning questions}

\begin{figure*}[t]
    \centering
    \includegraphics[width=\textwidth]
    {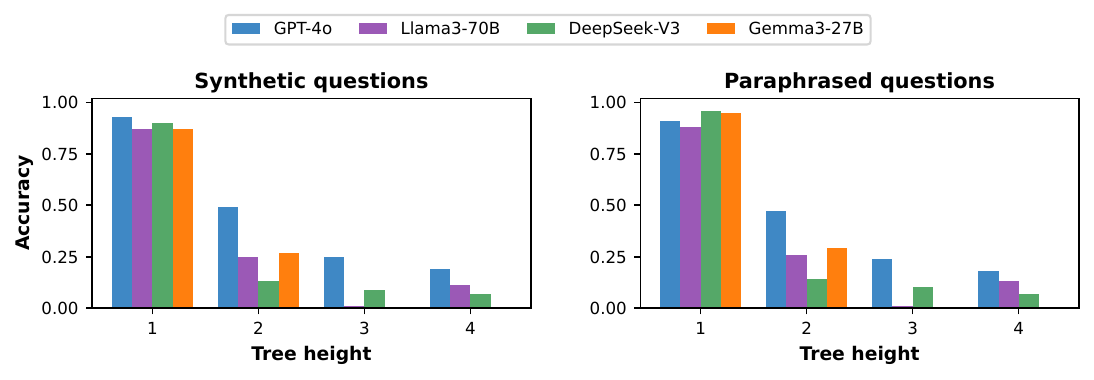}
    \caption{Accuracy on the original synthetic questions (left) and their
    natural-language paraphrases (right) across different ontology-tree
    heights.}
    \label{fig:paraphrased_results}
\end{figure*}

To determine whether failures arise from understanding paraphrased questions or solving logical reasoning problems, we evaluate each model on both original synthetic questions and their paraphrased versions. Figure~\ref{fig:paraphrased_results} shows only minor accuracy differences between the two formats. These results suggest that failures primarily stem from the underlying logical reasoning rather than understanding the paraphrased questions.

\section{Details of hypothesis quality metric}
\label{sec:metric}
\subsection{Theoretical justification} \label{sec:metric_theory}

\rev{Our quality metric can be motivated by an argument from Bayesian decision theory.
Let $T$ be a theory which consists of a set of observations
$o_1, \ldots, o_n$ and a set of proofs $\pi_1, \ldots, \pi_n$ where each
proof $\pi_i$ explains the observation $o_i$ (i.e., the conclusion of
$\pi_i$ is $o_i$).
Let
\[
p(T) = \prod_{i=1}^{n} p(\pi_i \mid \pi_1, \ldots, \pi_{i-1}) \, p(o_i \mid \pi_i)
\]
be a prior distribution over the theory.
Ideally, during inductive/abductive reasoning, given a new observation
$o_{n+1}$, we aim to find the proof $\pi_{n+1}$ that maximizes the posterior}

\rev{\begin{align*}
p(\pi_{n+1} \mid \pi_1,\ldots,\pi_n,o)
&\propto p(\pi_{n+1} \mid \pi_1,\ldots,\pi_n)  \\
&\quad \cdot p(o_{n+1} \mid \pi_{n+1})
\end{align*}}

\rev{Since
\[
p(o_i \mid \pi_i) = \mathds{1}\{o_i \text{ is the conclusion of } \pi_i\},
\]
the posterior is equal to
\[
p(\pi_{n+1} \mid \pi_1, \ldots, \pi_n)
\]
whenever the conclusion of $\pi_{n+1}$ is $o_{n+1}$.}

\rev{Now suppose we make the mild assumption that $p(T)$ is such that
$p(T) > p(T')$ whenever $T$ has fewer axioms in its proofs than does $T'$.
Therefore, $p(T)$ is maximized if $\pi_{n+1}$ minimally introduces new
axioms (and equivalently, maximally reuses existing axioms).
Note that the quality metric $g(\pi_{n+1})$ is the ratio of two quantities:
(1) the average number of times each premise $h \in \pi_{n+1}$ appears in
$\pi_1, \ldots, \pi_n$, and
(2) the average number of times each premise in the ground-truth proof
appears in $\pi_1, \ldots, \pi_n$.
We specifically selected each ground-truth proof to be the unique proof with the minimum number of
new axioms.
If we consider two proofs $\pi_{n+1}$ and $\pi'_{n+1}$ such that
$\pi_{n+1}$ has more new axioms than $\pi'_{n+1}$, then the sum in the
numerator of $g(\pi_{n+1})$ would contain more zero terms than that of
$g(\pi'_{n+1})$, and so
\[
g(\pi_{n+1}) < g(\pi'_{n+1})
\]
necessarily.}

\rev{Thus, with mild assumptions on the prior over the theory $p(T)$,
our proposed quality metric provides an easy-to-compute proxy for the
posterior probability of the theory, where more probable theories will
have higher quality scores. Furthermore, our quality metric is \emph{agnostic} to the specific choice of the prior $p(T)$.}

\subsection{Human validation experiments} \label{sec:metric_human_eval}
\rev{To evaluate whether the proposed metric aligns with human judgments of hypothesis quality, we conduct a human evaluation study. We recruit three independent annotators with graduate-level STEM degrees in US R1 universities. The evaluation dataset contains 100 reasoning questions, each associated with three candidate hypotheses. For each question, annotators are asked to select the hypothesis that better explains observations and is most concise. The presentation order of hypotheses is randomized to avoid positional bias. Each reasoning question is evaluated by three annotators, and the final preference label is determined by majority vote.}

\rev{To quantify alignment with human judgments, we measure agreement accuracy, defined as the percentage when the proposed metric assigns the highest score to the human-preferred hypothesis. The proposed metric achieves an agreement of 79\%, substantially outperforming a random baseline (33.3\%) and a length-based heuristic baseline (51\%). Human annotations exhibit substantial agreement (Fleiss’ \(\kappa\) = 0.75), indicating that the evaluation task is reasonably well-defined and that the proposed metric captures meaningful aspects of hypothesis quality.}

\rev{Please refer to Figure~\ref{fig:instruction_annotator} for the instruction to annotators.}

\begin{figure*}

\begin{tcolorbox}[
colback=gray!5,
colframe=black!60,
arc=3pt,
boxrule=0.6pt,
title=\textbf{Instruction for annotators}
]

\ttfamily

--- You will be given a set of reasoning questions. Each question includes:

* A description of observations

* Three candidate hypotheses that attempt to explain these observations

--- Your task is to select the single best hypothesis according to the following criteria:

   * The hypothesis should account for the given observations coherently and logically.

   * The hypothesis should be as concise as possible.

--- Important Guidelines

* Select only one hypothesis per question.

* Do not choose based on writing style, fluency, or formatting alone.

* Focus on the validity and quality of hypotheses, not superficial features.

* The order of hypotheses is randomized; treat each option independently.

* If multiple hypotheses seem similar, choose the one that is most concise.

--- Your selection will be used to determine the hypothesis that best explains the observations.

\normalfont
\end{tcolorbox}
\caption{Instruction for annotators.}

\label{fig:instruction_annotator}

\end{figure*}

\section{Algorithms}
\label{sec:algo}
Algorithms \ref{alg:hyp} and \ref{alg:cap} describe how we generate ontology trees and hypotheses for the examples in our framework.

\begin{algorithm}
\small
\let\oldnl\nl
\newcommand{\nonl}{\renewcommand{\nl}{\let\nl\oldnl}}
\SetNlSty{}{\color{Magenta}\sffamily}{}
\SetAlgoBlockMarkers{}{}
\SetKwProg{Fn}{function}{}{}
\SetKwIF{If}{ElseIf}{Else}{if}{ }{else if}{else }{}
\SetKw{Continue}{continue}
\SetKwFunction{FGenerateHypothesis}{generate\_hypothesis}
\SetKwFor{For}{for}{do}{end}
\SetKwFor{While}{while}{do}{end}
\SetKwProg{uForEach}{for each}{ do}{}
\SetKwProg{Fn}{function}{}{}
\AlgoDisplayBlockMarkers\SetAlgoVlined
\SetNoFillComment
\DontPrintSemicolon
\SetInd{0.0em}{0.8em}
    \Fn{\FGenerateHypothesis{ontology tree $T$}}{
        \uIf{task is ``\textit{infer property}''}{
            randomly select a member of the root node of $T$ \;
            randomly select a member of a leaf node of $T$ \;
            generate the observations where selected members have the property \;
        }\uElseIf{task is ``\textit{infer membership}''}{
            select a random leaf node of $T$ \;
            randomly select a property of the leaf node \;
            randomly select a property of the root node \;
            randomly select a property of the selected node's non-root ancestors \;
            generate observations where this entity has all properties \;
        }\ElseIf{task is ``\textit{infer subtype relation}''}{
            randomly select a member of the root node of $T$ \;
            randomly select a member of a leaf node of $T$ \;
            randomly select a member of a non-leaf of $T$ \;
            generate the observations where selected members also belong to a supertype \;
        }
        \Return{all world model axioms, observations, and the ground-truth hypothesis}
	}
	\caption{Generate a single hypothesis.}
	\label{alg:hyp}
\end{algorithm}


\begin{algorithm}
\small
\let\oldnl\nl
\newcommand{\nonl}{\renewcommand{\nl}{\let\nl\oldnl}}
\SetNlSty{}{\color{Magenta}\sffamily}{}
\SetAlgoBlockMarkers{}{}
\SetKwProg{Fn}{function}{}{}
\SetKwIF{If}{ElseIf}{Else}{if}{ }{else if}{else }{}
\SetKw{Continue}{continue}
\SetKwFunction{FGenerateExample}{generate\_example}
\SetKwFor{For}{for}{do}{end}
\SetKwFor{While}{while}{do}{end}
\SetKwFor{Repeat}{repeat}{}{end}
\SetKwProg{uForEach}{for each}{ do}{}
\SetKwProg{Fn}{function}{}{}
\AlgoDisplayBlockMarkers\SetAlgoVlined
\SetNoFillComment
\DontPrintSemicolon
\SetInd{0.0em}{0.8em}
    \Fn{\FGenerateExample{}}{
        \tcc*[f]{first, build the ontology tree} \;
        \tcc*[f]{\texttt{node()} initializes an empty node object} \;
        \tcc*[f]{add a root node} \;
        $tree\_nodes \gets$ [[\texttt{node()}]] \;
        \For{$i= 1,\hdots,h-1$}{
            $current\_nodes \gets []$ \;
            \For{$parent\_node \in tree\_nodes[-1]$}{
                $u \gets$ \texttt{random(0,1)} \;
                \uIf{$u > \frac{1}{2}$}{
                    \Repeat{2 times}{
                        $current\_nodes$\texttt{.append(node())} \;
                    }
                }\Else{
                    \Repeat{3 times}{
                        $current\_nodes$\texttt{.append(node())} \;
                    }
                }
            }
            $tree\_nodes$\texttt{.append($current\_nodes$)} \;
        }
        \For{$current\_nodes \in tree\_nodes$}{
            \For{$node \in current\_nodes$}{
                \tcc*[f]{allocate three members and properties} \;
                \Repeat{3 times}{
                    $node$\texttt{.members} \texttt{.append(} \texttt{random\_member())} \;
                    $node$\texttt{.properties} \texttt{.append(} \texttt{random\_property())} \;
                }
            }
        }
        \uIf{this is a single-hypothesis example}{
            randomly select a task type \;
            \texttt{generate\_hypothesis($tree\_nodes$)} \;
        }\lElse{
            generate 1 to 3 hypotheses per tree layer
        }
	}
	\caption{Generate an abductive/inductive reasoning example.}
	\label{alg:cap}
\end{algorithm}

\section{Results with absolute accuracy}
\label{sec:moreresults}
Here we list the absolute results of Figure \ref{fig:icl} in Figure \ref{fig:icl_absolute}.
\begin{figure*}
    \centering
    \resizebox{\textwidth}{!}{\input{figures/icl_absolute.pgf}}
        \caption{Absolute results of weak accuracy, strong accuracy, and quality versus the height of the ontology tree with multiple hypotheses under zero-shot (\textbf{top}), 8-shot in-distribution (\textbf{middle}), and 8-shot out-of-distribution (\textbf{bottom}).}
    \label{fig:icl_absolute}
\end{figure*}
\section{Reasoning-augmented prompts}
\label{sec:reasoning_augmented_prompts}
We provide representative examples of the chain-of-thought (CoT) and  bag-of-thoughts (BoT) prompts used in our experiments.

\paragraph{Chain-of-Thought (CoT).} Figure~\ref{fig:cot} presents an example of the CoT prompt used in our experiments.
\begin{figure}
\begin{tcolorbox}[
colback=gray!5,
colframe=black!60,
arc=3pt,
boxrule=0.6pt,
title=\textbf{Chain-of-Thought (CoT) Prompt}
]
\ttfamily
Amy is a dalpist. Pamela is a dalpist. Jerry is a dalpist. We observe that Amy is rainy. Jerry is rainy. Pamela is rainy. 

-

Please produce hypotheses to explain all observations. Let’s think step by step.
\normalfont
\end{tcolorbox}
\caption{An example of the chain-of-thought (CoT) prompt used in our experiments.}
\label{fig:cot}
\end{figure}

\paragraph{Bag-of-Thought (BoT).} Figure~\ref{fig:bot} presents an example of the BoT prompt used in our experiments.
\begin{figure}
\begin{tcolorbox}[
colback=gray!5,
colframe=black!60,
arc=3pt,
boxrule=0.6pt,
title=\textbf{Bag-of-Thought (BoT) Prompt}
]
\ttfamily
You are solving a problem using a reasoning template.

-

Template:

1. Construct the known ontology relationships.

2. Read observations carefully.

3. Come up with hypotheses in the ontology tree.

4. Use the hypotheses and the known ontology relationships to prove the observations.

-

Problem: Amy is a dalpist. Pamela is a dalpist. Jerry is a dalpist. We observe that Amy is rainy. Jerry is rainy. Pamela is rainy.

-

Please produce hypotheses to explain all observations. Follow the template to solve the problem.
\normalfont
\end{tcolorbox}
\caption{An example of the bag-of-thought (BoT) prompt used in our experiments.}
\label{fig:bot}
\end{figure}

\section{Error analysis}
\label{sec:response}
\subsection{Common error types and example responses}
\label{sec:examples}

In this section, we detail the most common types of errors that LLMs produce in our experiments. For each of the error types listed below, we provide an example consisting of a world model, list of observations, ground-truth hypotheses, and the mistaken LLM output. \\

\begin{tcolorbox}[
colback=gray!5,
colframe=black!60,
arc=3pt,
breakable,
boxrule=0.6pt,
left=4pt,right=4pt,
title=\textbf{Error Type 1:} \emph{Wrong ontology direction}]
\textbf{World Model:} Porpors are dalpists. Nicole is a gergit. Sarpers are dalpists. Every gergit is a dalpist. Debra is a dalpist. Each dalpist is rainy. Every porpor is not cold. Raymond is a gergit. Each porpor is translucent. Amy is a porpor. Jerry is a dalpist. Pamela is a dalpist. Sharon is a gergit. Edward is a dalpist. Michael is a sarper.
\\
\textbf{Observations:} Sharon is moderate. Barbara is not cold. Debra is liquid. Barbara is rainy. Barbara is liquid. Barbara is translucent. Amy is a rompus. Raymond is moderate. Jerry is a rompus. Edward is liquid. Pamela is a rompus. Nicole is moderate.
\\
\textbf{Ground-truth hypotheses:} Dalpists are rompuses. Every dalpist is liquid. Each gergit is moderate. Barbara is a porpor.
\tcblower
\textbf{Mistaken LLM Output:} \\ ... Rompuses are dalpists. ...
\end{tcolorbox}

\begin{tcolorbox}[
colback=gray!5,
colframe=black!60,
arc=3pt,
breakable,
boxrule=0.6pt,
left=4pt,right=4pt,
title=\textbf{Error Type 2:} \emph{Ignore the ontology and produce unnecessary hypotheses}
]
\textbf{World Model:} David is a sorple. Each sorple is a dropant. Each hilper is a thorpin. Rimpees are scrompists. Brian is a sorple. Dropants are gompers. Angela is a gomper. Each thorpin is a gomper. Every scrompist is a gomper.
\\
\textbf{Observations:} David is rainy. Brian is rainy. Angela is rainy.
\\
\textbf{Ground-truth hypotheses:} Every gomper is rainy.
\tcblower
\textbf{Mistaken LLM Output:} ... Every gomper is rainy. Every dropant is rainy. ...
\end{tcolorbox}

\begin{tcolorbox}[
colback=gray!5,
colframe=black!60,
arc=3pt,
breakable,
boxrule=0.6pt,
left=4pt,right=4pt,
title=\textbf{Error Type 3:} \emph{Fall back to rival hypotheses}
]
\textbf{World Model:} Rorpants are dolpers. Every worple is a wumpus. Each drompee is moderate. Ashley is a rorpant. Fimples are dolpers. Ronald is a wumpus. Wumpuses are gwompants. Yompins are dolpers. Brian is a dolper. Michelle is a worple. Each dolper is a gwompant. Shirley is a rorpant. Sandra is a bempin. Linda is a fimple. Drompees are pale. Jack is a yompin. Each drompee is a bempin. Bempins are cold. Bempins are gwompants. Emma is a gwompant. Michael is a gwompant.
\\
\textbf{Observations:} Shirley is a borpin. Christopher is pale. Emma is a borpin. Brian is salty. Michael is salty. Christopher is a borpin. Christopher is cold. Ashley is salty. Christopher is moderate.
\\
\textbf{Ground-truth hypotheses:} Each gwompant is a borpin. Each gwompant is salty. Christopher is a drompee.
\tcblower
\textbf{Mistaken LLM Output:} \\ ... Shirley is a borpin. ...
\end{tcolorbox}

\begin{tcolorbox}[
colback=gray!5,
colframe=black!60,
arc=3pt,
breakable,
boxrule=0.6pt,
left=4pt,right=4pt,
title=\textbf{Error Type 4:} \emph{Hallucinated entities}
]
\textbf{World Model:} Jonathan is a rifpist. Yimples are dull. Samantha is a rifpist. Karen is a gwompant. Scott is a jelgit. Olivia is a frompor. Maria is a lirpin. Twimpees are lirpins. Each rifpist is aggressive. Frompors are jelgits. Jessica is a jelgit. Amanda is a twimpee. Rifpists are jelgits. Each lirpin is a jelgit. Yimples are small. Each gwompant is a frompor. Each yimple is a rifpist.
\\
\textbf{Observations:} Mary is not overcast. Scott is not overcast. Jonathan is not overcast. Samantha is a kurpor. Mary is aggressive. Jessica is a kurpor. Olivia is a kurpor. Mary is small. Mary is dull.
\\
\textbf{Ground-truth hypotheses:} Every jelgit is a kurpor. Jelgits are not overcast. Mary is a yimple.
\tcblower
\textbf{Mistaken LLM Output:} \\ ... Mary is a dolphant. ...
\end{tcolorbox}

\subsection{Details of error analysis}
\begin{figure}[ht]
    \centering
    \resizebox{\textwidth}{!}{
    \includegraphics[width=\textwidth]{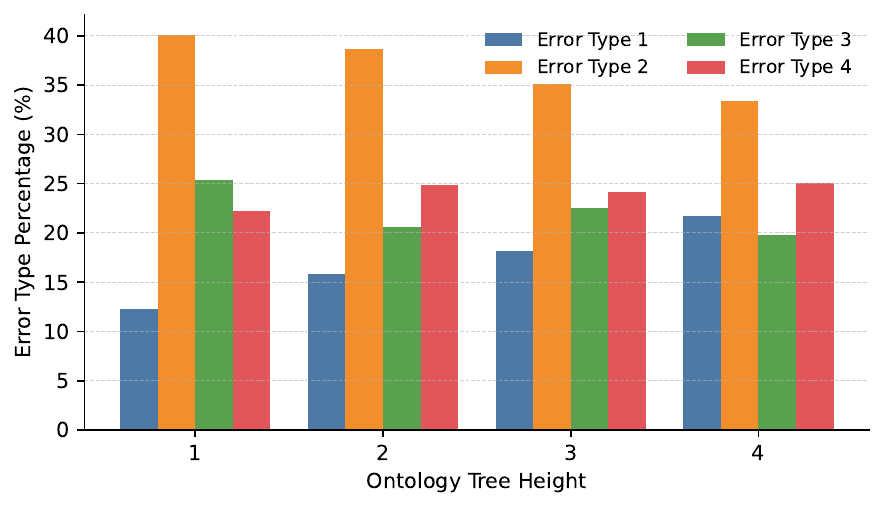}}
    \caption{Distribution of error types across ontology tree heights.}
    \label{fig:error_by_height}
\end{figure}

\rev{We first randomly sampled 200 LLM example responses to reasoning questions of varying difficulty. Then, we manually inspected them to identify recurring failure patterns as summarized in the paper. To further quantitatively study the error patterns of all responses, we use an LLM-as-a-judge \citep{llmasjudge} and prompt \textsc{GPT-4o} with in-context examples. We use the following prompt in Figure~\ref{fig:judge_prompt} for the LLM-as-a-judge. We categorize the error pattern distribution by ontology tree height and present the results in Figure~\ref{fig:error_by_height}. As shown in the figure, Error Type 2 (\emph{Ignore the ontology and produce unnecessary hypotheses}) is the most significant error across all ontology tree heights, indicating that LLMs can generate weakly correct hypotheses but fail to consider the given ontology tree and optimize them in the sense of Occam's razor.}


\begin{figure*}[t]

\begin{tcolorbox}[
colback=gray!5,
colframe=black!60,
arc=3pt,
boxrule=0.6pt,
title=\textbf{Prompt for Error Analysis}
]

\textbf{Prompt Template}

\ttfamily
You are an expert evaluator of reasoning quality. Your task is to analyze a model's reasoning process for a question and determine the type of error it contains.

\#\# Task

Given:

* A question \\
* The model’s reasoning process \\ 
* The correct answer

Identify which error type best describes the mistake in the reasoning.

Choose exactly one error type from the following categories.

---

\#\# Error Types

** Error Type 1: Wrong ontology direction**
The hypothesis contains the wrong ontology direction.

** Error Type 2: Fall back to trivial hypotheses**
The hypothesis is the observation itself.

** Error Type 3: Ignore the ontology and produce unnecessary hypotheses**
The hypothesis is technically correct but unnecessary given the ontology.

** Error Type 4: Hallucinated entities**
The reasoning relies on non-existent entities.

---

\#\# Examples

\#\#\# Example 1 (Wrong ontology direction)

<omitted for space, see \S\ref{sec:examples} for examples>

---

\#\#\# Example 2 (Fall back to trivial hypotheses)

<omitted for space, see \S\ref{sec:examples} for examples>

---

\#\#\# Example 3 (Ignore the ontology and produce unnecessary hypotheses)

<omitted for space, see \S\ref{sec:examples} for examples>

---

\#\#\# Example 4 (Hallucinated entities)

<omitted for space, see \S\ref{sec:examples} for examples>

---

\#\# Now evaluate the following case

Question:
\{\{QUESTION\}\}

Model Reasoning:
\{\{MODEL\_REASONING\}\}

Correct Answer:
\{\{CORRECT\_ANSWER\}\}

---

\#\# Output Format

Error Type: <one of: Wrong ontology direction | Fall back to trivial hypotheses | Ignore the ontology and produce unnecessary hypotheses | Hallucinated entities>

\normalfont
\end{tcolorbox}
\caption{Prompt used to analyze common error patterns.}

\label{fig:judge_prompt}

\end{figure*}

\label{sec:appendix}
\end{document}